\documentclass[letterpaper]{article} 
\usepackage{aaai25}  
\usepackage{times}  
\usepackage{helvet}  
\usepackage{courier}  
\usepackage[hyphens]{url}  
\usepackage{graphicx} 
\usepackage{natbib}  
\usepackage{caption} 
\usepackage{algorithm}
\usepackage{algorithmic}

\usepackage{caption}
\usepackage{subcaption}
\usepackage{multirow}
\usepackage{booktabs}
\usepackage{amsmath}

\usepackage{xspace}
\usepackage{bm}
\usepackage{amssymb}

\makeatletter
\DeclareRobustCommand\onedot{\futurelet\@let@token\@onedot}
\def\@onedot{\ifx\@let@token.\else.\null\fi\xspace}

\def\eg{\emph{e.g}\onedot} 
\def\ie{\emph{i.e}\onedot}

\makeatother

\usepackage{array}

\newlength\savewidth\newcommand\shline{\noalign{\global\savewidth\arrayrulewidth
  \global\arrayrulewidth 1pt}\hline\noalign{\global\arrayrulewidth\savewidth}}

\newcolumntype{x}[1]{>{\centering\arraybackslash}p{#1pt}}
\newcolumntype{y}[1]{>{\raggedright\arraybackslash}p{#1pt}}
\newcolumntype{z}[1]{>{\raggedleft\arraybackslash}p{#1pt}}

\usepackage{newfloat}
\usepackage{listings}
\DeclareCaptionStyle{ruled}{labelfont=normalfont,labelsep=colon,strut=off} 
\floatstyle{ruled}
\newfloat{listing}{tb}{lst}{}
\floatname{listing}{Listing}
\title{Wavelet-Driven Masked Image Modeling: A Path to 
Efficient \\ Visual Representation}
\author{
    Wenzhao Xiang\textsuperscript{1,2,3},
    Chang Liu\textsuperscript{4},
    Hongyang Yu\textsuperscript{2\thanks{Corresponding author.}},
    Xilin Chen\textsuperscript{1,3}
}
\affiliations{
    \textsuperscript{\rm 1}Key Laboratory of Intelligent Information Processing, Institute of Computing Technology, Chinese Academy of Sciences\\
     \textsuperscript{\rm 2}Peng Cheng Laboratory\\
    \textsuperscript{\rm 3} University of Chinese Academy of Sciences\\
    \textsuperscript{\rm 4} Department of Electronic Engineering, Shanghai Jiao Tong University \\

    {xiangwenzhao22@mails.ucas.ac.cn, sunrise6513@sjtu.edu.cn, yuhy01@pcl.ac.cn, xlchen@ict.ac.cn}
}

\usepackage{bibentry}

\begin{document}

\maketitle

\begin{abstract}
Masked Image Modeling (MIM) has garnered significant attention in self-supervised learning, thanks to its impressive capacity to learn scalable visual representations tailored for downstream tasks. However, images inherently contain abundant redundant information, leading the pixel-based MIM reconstruction process to focus excessively on finer details such as textures, thus prolonging training times unnecessarily. Addressing this challenge requires a shift towards a compact representation of features during MIM reconstruction. Frequency domain analysis provides a promising avenue for achieving compact image feature representation. In contrast to the commonly used Fourier transform, wavelet transform not only offers frequency information but also preserves spatial characteristics and multi-level features of the image. Additionally, the multi-level decomposition process of wavelet transformation aligns well with the hierarchical architecture of modern neural networks. In this study, we leverage wavelet transform as a tool for efficient representation learning to expedite the training process of MIM. Specifically, we conduct multi-level decomposition of images using wavelet transform, utilizing wavelet coefficients from different levels to construct distinct reconstruction targets representing various frequencies and scales. These reconstruction targets are then integrated into the MIM process, with adjustable weights assigned to prioritize the most crucial information. Extensive experiments demonstrate that our method achieves comparable or superior performance across various downstream tasks while exhibiting higher training efficiency.
\end{abstract}

%

\section{Introduction}
\label{sec:intro}
Driven by the success of masked language modeling (MLM) in natural language processing (NLP)~\cite{kenton2019bert,liu2019roberta,yang2019xlnet}, masked image modeling (MIM)~\cite{bao2021beit,he2022masked,xie2022simmim} has emerged as a leading approach in the realm of self-supervised visual representation learning. MIM helps the model learn scalable and rich representations for various downstream tasks by masking a portion of the input data and then predicting the missing data based on the visible parts.
As pixel-based MIM methods~\cite{he2022masked,xie2022simmim} have developed, researchers have begun directly employing raw pixel values as reconstruction targets, showcasing a simple and effective approach. 

\begin{figure}[tb]
  \centering
  \begin{subfigure}{0.56\linewidth}
    \begin{minipage}[b]{\linewidth}
        \includegraphics[width=\linewidth]{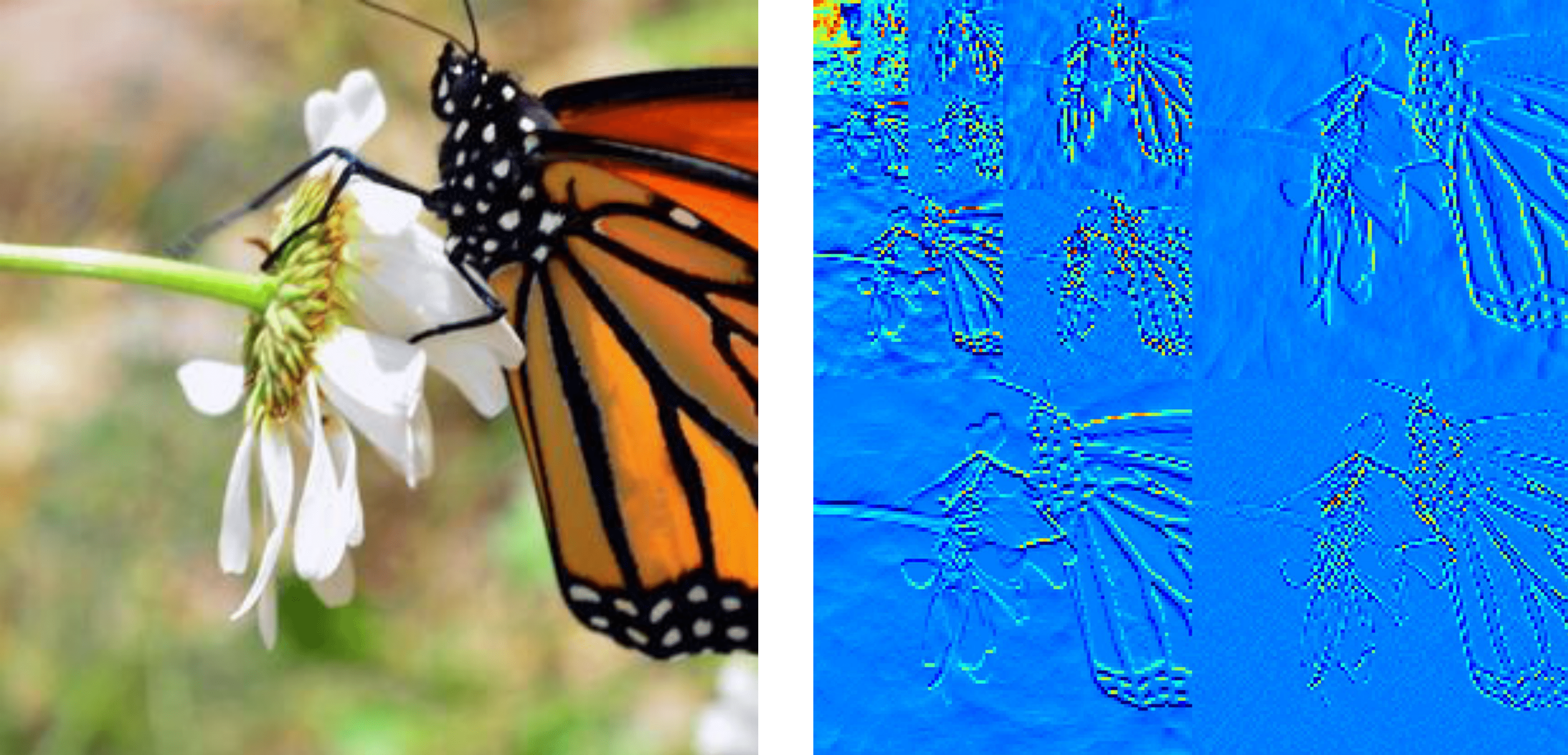}
        \caption{The Visualization of wavelet coefficients obtained from multi-level wavelet decomposition of an image.}
        \label{fig:waveletcoffi}
    \end{minipage}
    \begin{minipage}[b]{\linewidth}
        \includegraphics[width=\linewidth]{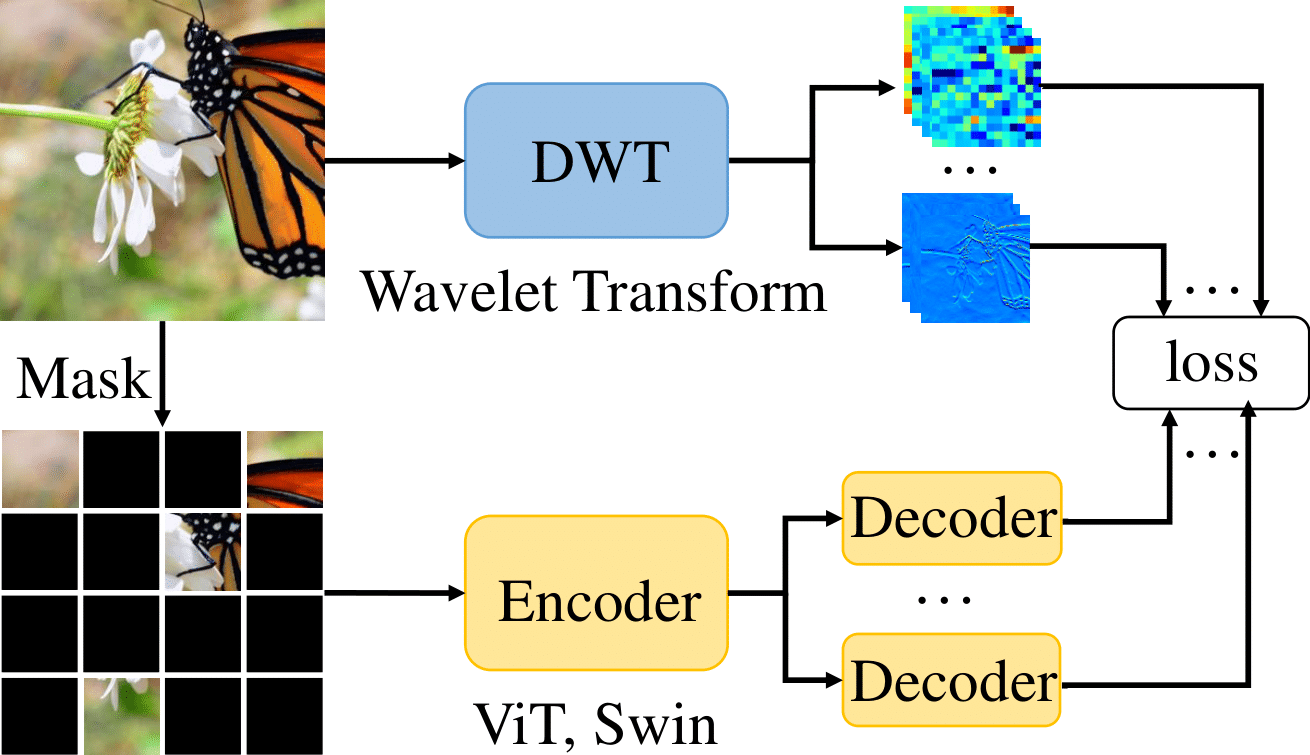}
        \caption{The simplified pipeline of our proposed framework.}
        \label{fig:pipeline}
    \end{minipage}
  \end{subfigure}
  \begin{subfigure}{0.405\linewidth}
    \includegraphics[width=\linewidth]{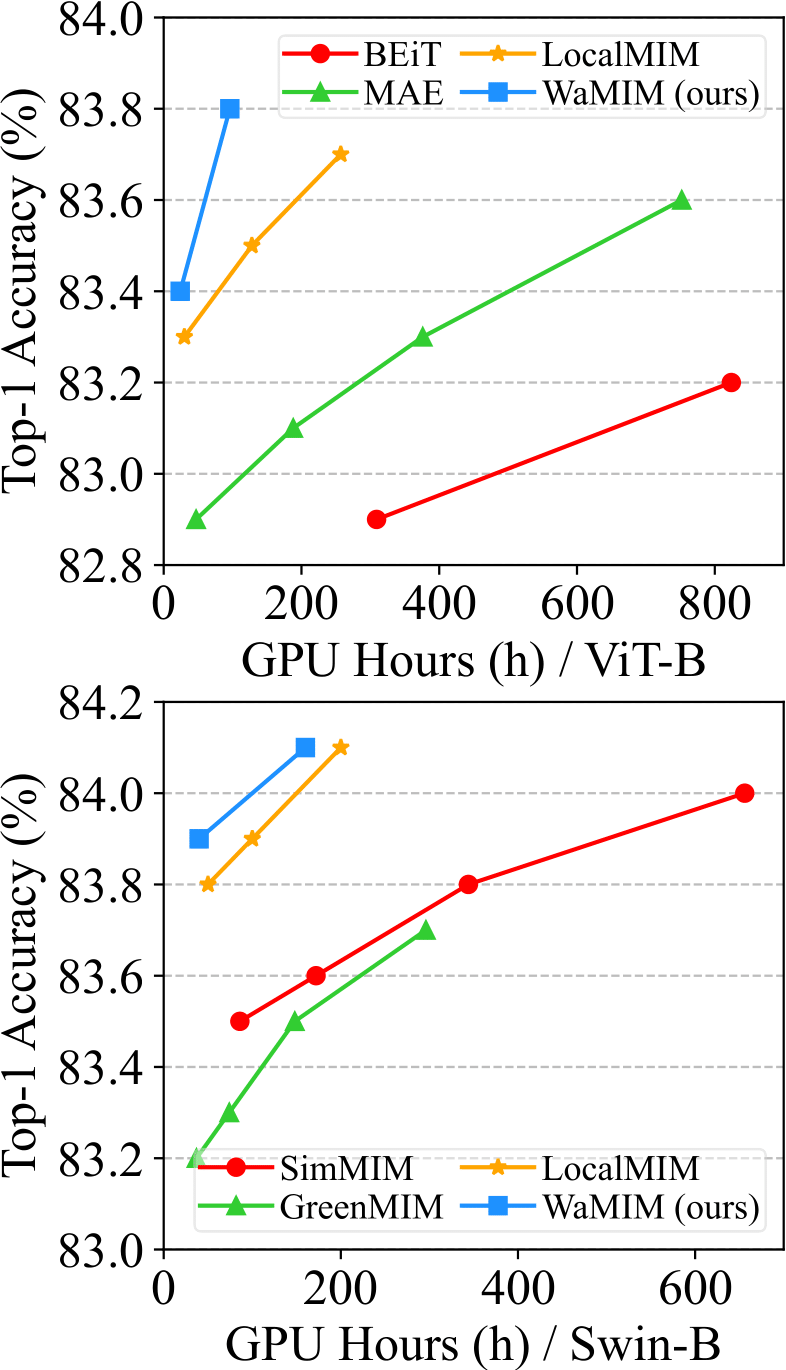}
    \caption{The Top-1 fine-tuning accuracy on ImageNet-1K vs. 'GPU Hours'. 
    }
    \label{fig:performance}
  \end{subfigure}
  \caption{An overview of the proposed WaMIM.}
  \label{fig:overview}
\end{figure}

While using raw pixel values as reconstruction targets preserves the original information in MIM, the raw pixels of an image often contain a substantial amount of redundant information that isn't particularly beneficial for representation learning. Some studies~\cite{liu2023pixmim,liu2023improving} suggest that directly predicting raw pixel values can make pre-training overly sensitive to fine-grained details of images, such as textures. Consequently, this can render the self-supervised pre-training process more challenging and unstable, requiring prolonged training to acquire rich and effective representations for downstream tasks. This necessitates a discriminative and compact feature representation for the reconstruction targets in the MIM process.

Some studies~\cite{xie2022masked,liu2023pixmim,liu2023improving} have tackled these issues by shifting their focus to the frequency domain. They achieve this by partitioning frequency bands and utilizing more specific low-frequency or high-frequency information to construct compact reconstruction targets. These methods commonly employ the Fourier transform for frequency domain analysis. However, the Fourier transform only provides global frequency characteristics of signals while completely disregarding their local spatial properties, leading to the loss of valuable information. Compared to the Fourier transform, the wavelet transform preserves the spatial properties of an image while providing its frequency characteristics~\cite{arts2022fast}. Additionally, the multi-level decomposition mechanism helps extract multi-scale features of the image, which are highly beneficial in enhancing both the speed and effectiveness of MIM~\cite{wang2023masked, ren2023deepmim}. 

In this paper, we introduce an efficient MIM training framework, leveraging wavelet transform to build the reconstruction targets from both frequency and multi-scale perspectives. We first perform wavelet multi-level decomposition on the input image, obtaining multi-level wavelet coefficients representing different frequency bands and scales. Then, a subset of these wavelet coefficients is selected to construct the multi-level reconstruction targets. 
Given that wavelet coefficients inherently act as feature descriptors reflecting features of different frequencies and scales within the image~\cite{7844359}, as depicted in Figure~\ref{fig:waveletcoffi}, our framework directly employs these coefficients as multi-level reconstruction targets.
The output features from different encoder layers are decoded to predict the targets, with the shallow layers corresponding to high-frequency information and deep layers to low-frequency information.
This strategy is based on the observation that shallow layers of neural networks tend to capture low-level details of images, while deeper layers tend to capture high-level semantics~\cite{park2022vision,gao2022convmae}. 
The loss weights of different reconstruction targets can be adjusted to prioritize more crucial features during pre-training, thereby accelerating the representation learning process.
We denote the proposed method as "\textbf{Wa}velet-Driven \textbf{M}asked \textbf{I}mage \textbf{M}odeling" (WaMIM), as illustrated in Figure~\ref{fig:overview}. 

Overall, our contributions can be summarized as follows:

\noindent \ \textbullet \ Our approach introduces a framework to accelerate and enhance MIM pre-training by incorporating multi-level reconstruction targets, generated with the guidance of wavelet transform in both spatial and frequency domains. To the best of our knowledge, we are the first to introduce wavelet guidance into the MIM pre-training.

\noindent \ \textbullet \ We propose to directly utilize the wavelet coefficients as the multi-level reconstruction targets, which can be effortlessly integrated into most existing MIM frameworks.

\noindent \ \textbullet \ Extensive experiments demonstrate that our method can accelerate and enhance the MIM methods, as shown in Figure~\ref{fig:performance} and Table~\ref{tab:imagenet}. For example, WaMIM achieves a fine-tuning accuracy of 82.0\%/83.8\% on ViT-S/B with only 45/96 GPU hours, reducing the computational cost to just 47\%/13\% of MAE while demonstrating similar or slightly improved performance(1.1\%/0.2\% increase). Similar experimental results can be observed on the Swin Transformer and other downstream tasks.

\section{Related Work}

\noindent\textbf{Masked Image Modeling.} 
Building on the success of masked language modeling (MLM), particularly with models like BERT~\cite{kenton2019bert}, masked image modeling (MIM) has emerged as a parallel approach in the visual domain, learning robust visual representations by predicting masked image regions. BEiT~\cite{bao2021beit} is one of the pioneering methods in this space, utilizing an offline tokenizer, VQ-VAE~\cite{van2017neural}, to convert images into discrete visual tokens and establish a patch-level dictionary, to recover the token IDs for the masked patches.
iBOT~\cite{zhou2022image} takes a different approach by employing an online tokenizer, which generates targets for the encoder in a self-distillation framework for pre-training. Other methods, such as MAE~\cite{he2022masked} and SimMIM~\cite{xie2022simmim}, shift the focus to directly reconstructing raw pixel values, with MAE concentrating on visible patches and SimMIM on all patches. MaskFeat~\cite{wei2022masked}, on the other hand, avoids predicting raw pixel values by using low-level features like Histograms of Oriented Gradients (HOG)~\cite{dalal2005histograms} as reconstruction targets.
In addition to these, some approaches~\cite{wang2023masked,ren2023deepmim} explore the use of deep features, frequencies, or introduce novel pretext tasks like reconstructing corrupted images~\cite{fang2022corrupted} or denoising images~\cite{you2024beyond, xiang2024aemim}. More recent research~\cite {wang2023masked,ren2023deepmim} has proposed integrating multi-level supervision mechanisms into MIM, leading to significant improvements in both the speed and effectiveness of representation learning.

\noindent\textbf{Frequency Domain Analysis.} Frequency domain analysis has been extensively utilized across numerous computer vision tasks, including representation learning \cite{
xu2020learning,li2023discrete,liu2023improving2,zhu2024wavelet}, image generation \cite{jiang2021focal,phung2023wavelet}, and image super-resolution \cite{fuoli2021fourier,liu2023spectral}. For a long time, frequency analysis, encompassing Fourier analysis and wavelet analysis, has served as a pivotal tool in handling signals due to its inherent capability to segregate information across various scales. The efficacy of frequency analysis in expediting the training process of deep neural networks has been corroborated, as demonstrated in previous works such as \cite{yao2022wave, park2023rgb}. The efficiency of training MIM models is also dependent on a more effective and dense representation of image information. Besides, some findings indicate that shallow layers in MAE encompass notably more high-frequency components compared to deep layers, which primarily capture low-level details such as textures \cite{liu2023improving}. This insight suggests the potential of generating targets based on information in the frequency domain. Several approaches \cite{xie2022masked, liu2023devil, liu2023pixmim} leverage frequency analysis to extract crucial features for reconstruction targets.

\begin{figure*}[tb]
  \centering
  \includegraphics[height=7.5cm]{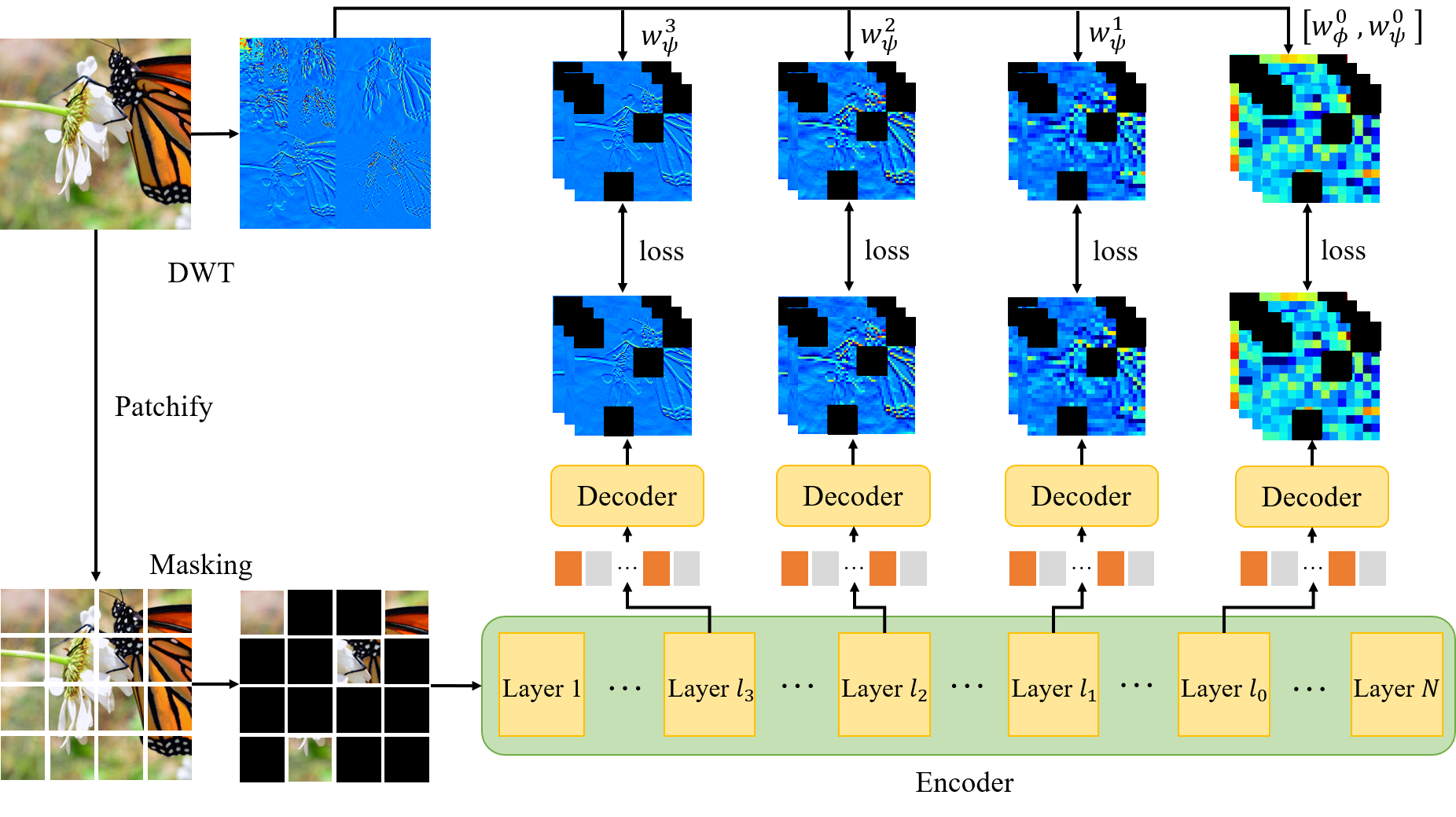}
  \caption{
  The training framework of WaMIM. The images are first patchified and masked. They are then passed through an encoder to extract features from various layers. The decoder selects features from specific layers to reconstruct the targets, which are generated using discrete wavelet transform (DWT) to decompose the images into different scales. The wavelet coefficient $\bm{w}^{K-k}_{\psi}$ serves as the reconstruction target for features extracted from Layer $l_{k}$. The larger $k$ corresponds to deeper layers in the encoder, aim at reconstructing higher frequency wavelet coefficients. $K$ represents the total number of layers selected for reconstruction. For $k=K$, the coefficient of the scaling function is concatenated with that of the wavelet function.
  }
  \label{fig:framework}
\end{figure*}

\section{Proposed Method}
\subsection{Preliminaries}

\noindent\textbf{Masked Image Modeling.}
In the MIM approach, models are tasked with predicting information from images with certain regions masked. To achieve this objective, the overarching framework comprises three main components: image masking, an auto-encoder model, and a target generator.

Formally, let $\bm{x} \in \mathbb{R}^{H \times W \times C}$ denote the input image, where $H$, $W$, and $C$ represent the height, width, and number of channels of the image respectively. The input images undergo masking with $f_{m}$, denoted as $\hat{\bm{x}}=f_{m}(\bm{x})$. The masking operation reorganizes the pixels into $N=HW//p^2$ non-overlapping patches, where $p$ represents the patch size, and masks the patches with a random mask $m \in \{0,1\}^N$, where $m_i=1$ indicates the patch is masked.

The corrupted image is then fed into an auto-encoder $f_{AE}$ to predict the target information of the original image. Typically, the auto-encoder comprises an encoder $f_{EN}$ and a decoder $f_{DE}$. Assuming the encoder has $L$ layers, the output of the encoder contains features from different scales, i.e., $[\bm{z}^{1},\bm{z}^{2},\cdots,\bm{z}^{L}]=f_{EN}(\hat{\bm{x}})$, and the output of the last layer of the encoder $\bm{z}^{L}=f_{EN}^{L}(\hat{\bm{x}})$ is decoded as $\hat{\bm{y}}^{L}=f_{DE}(\bm{z}^{L})$, representing the prediction for the target $\bm{y}$. This concept is further extended in \cite{wang2023masked}, which suggests that local multi-scale reconstruction offers a more efficient approach. Consequently, a more comprehensive objective for MIM involves reconstructing information from multiple layers. For each latent code $\bm{z}^l$, where $l\in[1,2,\cdots, L]$, the information is reconstructed as $\hat{\bm{y}}^l=f_{DE}^l(\bm{z}^l)$, with $\bm{y}^l$ representing the target. The reconstruction target is crucial in MIM and is generated by a target generator function $f_t$, such that $\bm{y}^l=f_t^l(\bm{x})$. Combining these elements yields the general formulation for MIM as
\begin{equation}
\label{eq:mim}
    \mathcal{L}=\sum_{l=1}^L c^l \cdot \mathcal{D}(f_{AE}^l(f_m(\bm{x})),f_t^l(\bm{x})),
\end{equation}
in which, $\mathcal{D}$ measures the distance (\eg, L1 or L2 distance) between the prediction and the target, $f_{AE}^l=f_{DE}^l \circ f_{EN}^l$, $c^l$ is the weight for the reconstruction loss of each layer. It should be noted that the distance measurement $\mathcal{D}(\cdot,\cdot)$ implicitly contains the operation of the masking operation to accelerate calculation. From this perspective, MAE \cite{he2022masked} is a special case of MIM with $c^l=1$ for $l = L$, $c^l=0$ for $l\neq L$ and $f_t^l(\bm{x})=\bm{x}$.

\noindent\textbf{Wavelet Transform.} Different from Fourier analysis, which examines signals solely in the frequency domain, wavelet analysis strikes a balance between time-domain and frequency-domain analysis. The wavelet transform decomposes signals using a dilated and translated version of a scaling function $\phi(t)$ and a wavelet function $\psi(t)$, both of which are defined as signals with zero mean and finite energy. In this paper, our focus lies on 2D discrete wavelet transform (DWT). Here, the scaling function is expanded as $\phi(t_m,t_n)$, and the wavelet function is expanded into three two-dimensional functions: $\psi^H(t_m,t_n)$ (measuring variations along columns), $\psi^V(t_m,t_n)$ (measuring variations along rows), and $\psi^D(t_m,t_n)$ (measuring variations along diagonals). Utilizing these scaling and wavelet functions, we define the scaled and translated basis functions as
\begin{gather}
    \phi_{j,m,n}(t_m,t_n)=2^{\frac{j}{2}}\phi(2^j t_m-m,2^j t_n-n),\\
    \psi_{j,m,n}^i(t_m,t_n)=2^{\frac{j}{2}}\psi^i(2^j t_m-m,2^j t_n-n), i={H,V,D},
\end{gather}
where index $i$ identifies the direction of the wavelet function.

The space spanned by $\phi_{j,m,n}$ is nested, with $j$ increasing, resulting in the entire space being defined within $L^2(\mathbb{R})$. The space spanned by $\psi_{j,m,n}^i$ is orthogonal to the space spanned by $\phi_{j,m,n}$ at the same $j$, ensuring that the combination of ${\phi_{j_0,m,n}, \psi_{j_0,m,n}^i, \cdots, \psi_{j_\infty,m,n}^i}$ forms the basis to represent two-dimensional signals. For an image $\bm{x}$ with size $M\times N$, we typically set $j_0=0$ and choose $N=M=2^J$, so that $j=0,1,\cdots,J$ and $m=n=0,1,\cdots,2^j-1$. The wavelet transform can be expressed as
\begin{equation}
\label{eq:wt}
    [\bm{w}_\phi^{0}; \bm{w}_\psi^{0},\cdots,\bm{w}_\psi^{J}]=f_w(\bm{x}),
\end{equation}
where $f_w$ is the forward function of DWT. Detailed expression is shown as
\begin{align}
\label{eq:5}
    \bm{w}_\phi^{0}[m,n]=\frac{1}{\sqrt{MN}}\sum_{t_m=0}^{M-1} \sum_{t_n=0}^{N-1} p(t_m,t_n) \phi_{0,m,n}(t_m,t_n), \\ \nonumber
    \bm{w}_\psi^{j}[i,m,n]=\frac{1}{\sqrt{MN}}\sum_{t_m=0}^{M-1} \sum_{t_n=0}^{N-1} p(t_m,t_n) \psi^i_{j,m,n}(t_m,t_n),
\end{align}
where $i=\{H,V,D\}$, $p(t_m,t_n)$ is the pixel value function of $\bm{x}$. The expression provided above is a common representation of the DWT. In practice, DWT is often implemented using Multi-Resolution Analysis, which facilitates a fast algorithm and eliminates redundant information.

\subsection{Wavelet-Driven Reconstruction Target}
\label{sec:3.2}
Creating reconstruction targets is a critical component of the MIM method. Effective reconstruction targets not only aid the model in acquiring superior representations for downstream tasks but also enhance pre-training efficiency and conserve computational resources. Following the works of MAE and SimMIM, many methods have employed raw or normalized pixel values as reconstruction targets. However, raw pixel values contain a considerable amount of redundant information. Directly employing them as reconstruction targets may result in the reconstruction task being overly sensitive to the high-frequency components of the image, which could be detrimental to representation learning.

Using frequency analysis tools can address the mentioned issue by selectively processing various frequency bands of the reconstruction targets. The Fourier transform is the most common tool for frequency domain analysis as in \cite{xie2022masked, liu2023devil, liu2023pixmim}. However, it only offers the global frequency response of the signal. Relying solely on the Fourier transform may result in the loss of local spatial characteristics and multi-scale information in the input image, which are also critical factors in enhancing representation learning~\cite{wang2023masked,ren2023deepmim}. 

In our method, we utilize the wavelet transform to guide the construction of reconstruction targets. 
In contrast to the Fourier transform, the wavelet transform, as shown in Equation~\ref{eq:wt}, maintains both spatial and frequency characteristics of an image, with its multi-level decomposition mechanism facilitating the extraction of multi-scale features of the image. As the wavelet coefficients are commonly used as the feature descriptor of input signals, for simplicity, we directly set the wavelet coefficients from DWT as the targets of the generator $f_t$. As $j$ in Equation~\ref{eq:5} increases, the coefficients refer to information with higher frequency, which are related to low-level features, \ie, features extracted from the shallow layer of the encoder. Therefore, we select $K$ ($K=J+1, K\leq L$) sets of features extracted from layers $\{l_1, l_2, \cdots, l_K\} \subseteq \{1,2, \cdots, L\}$. Then, we set the target for the features encoded from the $l_k$-th layer ($k \in \{1,\cdots, K\}$) as the wavelet coefficients of $j=K-k$, which is shown as
\begin{equation}
    f_t^{l_k}(\bm{x})=\begin{cases}
        \bm{w}^{K-k}_{\psi}, &\text{if}\ k\neq K \\
        [\bm{w}^{K-k}_{\phi}, \bm{w}^{K-k}_{\psi}], &\text{if}\  k= K,
    \end{cases}
\end{equation}
in which, $[\cdot,\cdot]$ is concatenation operation. It should be noted that the size for $\bm{w}^j_\phi$ is down-sampled as $j$ increases during the fast DWT process. Besides, $k$ is negative related to $j$, which means features from shallow layers (smaller $k$) correspond to coefficients of higher frequency (larger $j$). In this way, the target for MIM becomes reconstructing the wavelet coefficients of frequency $j=K-k$ with the features from layer $l_k$ of the encoder.

\subsection{Implementation Details}
Our approach, called Wavelet-Driven Masked Image Modeling (WaMIM), is a simple yet powerful method for self-supervised pre-training. We leverage the wavelet transform tool to create multiple reconstruction targets representing different frequencies and scales. And multi-level features from different layers are utilized to predict these targets separately, ensuring that the model comprehensively learns rich image representations. Figure~\ref{fig:framework} shows the overview of our framework. 
We then sequentially detail the key components of our framework, including the encoder, decoder, masking strategy, reconstruction target, and training objectives.

\noindent\textbf{Encoder.} We adopt vision transformer and swin transformer as the foundational architectures for the encoder $f_{EN}$. The input image $\bm{x}$ is first divided into regular non-overlapping patches, then some patches are masked out from the input, yielding a masked image $\hat{\bm{x}}$. For the vision transformer, only visible patches are fed into the encoder for efficiency. Thanks to the optimizations introduced by GreenMIM~\cite{huang2022green}, we apply the same mechanism to the swin transformer as well. The input patches are embedded by a linear projection layer, incorporating positional embeddings, and then processed via a series of self-attention-based Transformer blocks. We sequentially select $K$ features $\{\bm{z}^{l_k}|k\in \{1,2,\cdots,K\}\}$, which comes from shallow to deep layers as the multi-level features outputted by the encoder. $K$ equals the number of reconstruction targets. The encoder is further used for the various downstream tasks. 

\noindent\textbf{Decoder.} We employ multiple decoders to predict different reconstruction targets using the encoded multi-level features. The decoders take the output features of the encoder along with mask tokens as input, where the mask token represents a shared and learnable vector. To mitigate the computational overhead resulting from the increase in decoder numbers and improve training efficiency, we adopt light-weight decoders, consisting of only one transformer block and a linear layer, for example. We denote the decoder attached to the feature of the $l$-th layer as $f_{DE}^l$.

\noindent\textbf{Masking Strategy.} We adopt the random block-wise masking strategy in our framework. Given that our method incorporates multiple reconstruction targets across various scales, it's imperative to rescale the mask accordingly to match the corresponding scale of the reconstruction target during the computation of the reconstruction loss.

\noindent\textbf{Reconstruction Targets.}
As shown in Section~\ref{sec:3.2}, we first perform multi-level wavelet decomposition on the input image to obtain multi-level wavelet coefficients. These wavelet coefficients are further utilized as multi-level reconstruction targets representing different frequencies and scales. 
For transformers, shallow layers generally learn low-level representations of images, \eg textures and edges, while deeper layers capture higher-level semantic information, \eg object shapes. Therefore, in our framework, the shallow-layer output features of the encoder are employed to predict high-frequency (low-level) reconstruction targets, whereas deep or final-layer output features are used to predict low-frequency (high-level) reconstruction targets. We denote the reconstruction target for $l$-th layer as $f_t^l(\bm{x})$. 

\noindent\textbf{Training objectives.} 
The final reconstruction loss is defined by replacing the target generator in Equation~\ref{eq:mim} with wavelet coefficients at different scales, as expressed in
\begin{equation}
    \begin{aligned}
    \mathcal{L}&=\sum_{k=1}^{K-1} c^{l_k} \mathcal{D}(f_{AE}^{l_k}(f_m(\bm{x})),\bm{w}^{K-k}_\phi) \\
    &+c^{l_K} \cdot \mathcal{D}(f_{AE}^{l_K}(f_m(\bm{x})),[\bm{w}^0_\psi,\bm{w}^0_\phi]),
    \end{aligned}
\end{equation}
where outputs of $f_{AE}^{l_k}$ will be made adaptive to different sizes of the target. Besides, the distance measurement $\mathcal{D}(\cdot,\cdot)$ encompasses masking operation considering different scales of the wavelet coefficients.

\section{Experiments}
\subsection{Experimental settings}
\noindent\textbf{Pre-training setup.} 
We perform pre-training on the ImageNet-1K dataset~\cite{russakovsky2015imagenet} without any ground-truth labels. We use the columnar ViT~\cite{dosovitskiy2021an} and pyramidal Swin~\cite{liu2021swin} architectures for the encoder, with an input size of $224 \times 224$. The input images are segmented into patches of size $p = 16$ for ViT and $p = 4$ for Swin, and are randomly masked with a default ratio of $r = 0.75$. Basic data augmentations, including random cropping and horizontal flipping, are applied. For each architecture, we construct four reconstruction targets that vary from high-frequency, low-level to low-frequency, high-level. We perform a 5-level wavelet decomposition on the input image and select wavelet coefficients from levels $2$ to $5$ as the reconstruction targets, resulting in scales of $\{56^2, 28^2, 14^2, 7^2\}$. In the Swin architecture, we use output features from stages $\{2, 4, 22, 24\}$ for prediction. Each feature's decoder consists of a transformer block with an embedding dimension of 128 and 4 attention heads. For ViT, the chosen layers are $\{3, 6, 9, 12\}$, with each decoder comprising a transformer block with an embedding dimension of 256 and 8 attention heads. Loss weights are set to $\{0.8, 0.9, 1.1, 1.2\}$. Haar wavelet basis is employed for wavelet transform, and the wavelet coefficients used for reconstruction are normalized. Detailed pre-training procedures are provided in the appendix.

\noindent\textbf{Image Classification.} After pre-training, we first conduct supervised end-to-end fine-tuning on ImageNet-1K~\cite{russakovsky2015imagenet} dataset. 
The fine-tuning of all models is conducted using images with resolutions of $224 \times 224$. The full fine-tuning details can be found in the appendix.

\noindent\textbf{Object Detection and Instance Segmentation.} 
We adopt the pre-trained Swin models as the feature extractor for Mask R-CNN ~\cite{he2017mask}. We start by fine-tuning Mask R-CNN with the COCO~\cite{lin2014microsoft} train2017 dataset split, followed by an evaluation of its performance on the val2017 split using AP$^{b}$ and AP$^{m}$ metrics. The fine-tuning process is carried out using a $3 \times$ schedule, consisting of 36 training epochs. We follow the implementation of Mask R-CNN provided by MMDetection~\cite{chen2019mmdetection}. The full fine-tuning details are shown in the appendix.

\noindent\textbf{Semantic Segmentation.} 
We adapt the pre-trained ViT models for semantic segmentation using UperNet~\cite{xiao2018unified} as the segmentor. We perform end-to-end fine-tuning on the ADE20k dataset~\cite{zhou2017scene} for $160k$ iterations with an input resolution of $512 \times 512$, and evaluate performance on the validation set using the mIoU metric. Full fine-tuning details are provided in the appendix.

\subsection{Main Results}
\begin{table*}[tb]
  \centering
  \begin{tabular}{@{}lcccccc@{}}
    \hline
    Method & Model & Target & PT Epoch& GPU Hrs./Ep.& Total GPU Hrs. &  Acc.\\
    \hline
    MAE*~\cite{he2022masked}  & ViT-S & Pixel&300&0.32&96&80.9\\
    LocalMIM*~\cite{wang2023masked} & ViT-S & Hog& 300& 0.20&60&81.6\\
    MFM~\cite{xie2022masked} & ViT-S & Fourier &300&0.60 & 180&81.6\\
    WaMIM & ViT-S & Wavelet &100&0.15 &15 &81.8\\
    WaMIM & ViT-S & Wavelet &300& 0.15&45 &\textbf{82.0}\\
    \hline
    MoCo v3~\cite{chen2021empirical} & ViT-B & Momentum & 600&- &- &83.2\\
    iBOT~\cite{zhou2022image} & ViT-B & Momentum &400&4.33&1732&83.8\\
    BEiT~\cite{bao2021beit} & ViT-B & DALLE &800&1.03&824&83.2\\
    MAE*~\cite{he2022masked}  & ViT-B & Pixel&800&0.47&376&83.3\\
    MAE~\cite{he2022masked}  & ViT-B & Pixel&1600&0.47&752&83.6\\
    MaskFeat\cite{wei2022masked} & ViT-B & Hog&1600&1.67&2672&\textbf{84.0}\\
    CAE~\cite{chen2024context} & ViT-B & DALLE&800&1.20&960&83.6\\
    LocalMIM~\cite{wang2023masked} & ViT-B & Hog& 100& 0.30&30&83.3\\
    LocalMIM*~\cite{wang2023masked} & ViT-B & Hog& 400& 0.30& 120&83.5\\
    MFM~\cite{xie2022masked} & ViT-B & Fourier& 300&1.10 &330 &83.1\\
    WaMIM & ViT-B & Wavelet&100&0.24& 24 &83.4\\
    WaMIM & ViT-B & Wavelet&400&0.24& 96 &83.8\\
    \hline
    SimMIM$_{192}$~\cite{xie2022simmim} & Swin-B & Pixel&800&0.82&656&84.0\\
    GreenMIM~\cite{huang2022green} & Swin-B & Pixel&800&0.37&296&83.7\\
    LocalMIM~\cite{wang2023masked} & Swin-B & Hog&100&0.50&50&83.8\\
    LocalMIM~\cite{wang2023masked} & Swin-B & Hog&400&0.50&200&\textbf{84.1}\\
    WaMIM & Swin-B & Wavelet&100&0.40&40&83.9\\
    WaMIM & Swin-B & Wavelet&400&0.40&160&\textbf{84.1}\\
  \hline
  \end{tabular}
  \caption{
  Top-1 accuracy (\%) on ImageNet-1K. All models are pre-trained and fine-tuned at a resolution of 224 × 224, except for SimMIM$_{192}$, which uses a 192 × 192 resolution during pre-training. * indicates results reproduced using the official code.
  }
  \label{tab:imagenet}
  \vspace{-0.5em}
\end{table*}
\noindent\textbf{Image Classification.} We evaluate our WaMIM against existing MIM models, analyzing both pre-training efficiency and top-1 fine-tuning accuracy. The results are illustrated in Figure~\ref{fig:performance} and summarized in Table~\ref{tab:imagenet}. To ensure a fair comparison, we compute the pre-training efficiency of each method on identical hardware, utilizing a single Tesla A100-40G GPU, CUDA 11.7, and PyTorch 1.13.

For ViT-B, our method achieves $83.8\%$ accuracy with only 400 epochs, $96$ total GPU hours, which is $0.2\%$ higher than the MAE at 1600 epochs while requiring only $13\%$ of the computational cost. Regarding MFM, another method that employs the Fourier transform as the frequency analysis tool, our approach achieves a $0.7\%$ higher fine-tuning accuracy while incurring only a $29\%$ computational cost. This further highlights the superiority of wavelet transform over Fourier transform in MIM methods. When compared to LocalMIM, another multi-scale MIM method, our approach can also achieve better performance at both 100 epochs and 400 epochs, with higher training efficiency.

For the small-scale model ViT-S, our method can significantly enhance the performance even further.
Our method at 100 epochs achieves a fine-tuning accuracy of $81.8\%$, which is $0.9\%$, $0.2\%$, and $0.2\%$ higher than MAE, MFM, and LocalMIM at 300 epochs, despite using only $16\%$, $8\%$, and $25\%$ computational costs. When enlarging the training schedule to 300 epochs, our method achieves an accuracy of $82.0\%$, marking a new state-of-the-art in terms of balanced performance and efficiency, to the best of our knowledge. 

For the hierarchical Swin-B architecture, our method achieves a fine-tuning accuracy of $83.9\%$ with only 100 epochs of pre-training, significantly outperforming SimMIM and GreenMIM in efficiency. Compared to LocalMIM, our method yields a slightly higher fine-tuning accuracy of $0.1\%$ at 100 epochs, while requiring just $80\%$ of the computational cost. With an extended training schedule of 400 epochs, our method reaches an accuracy of $84.1\%$, solidifying its position as the top performer in both performance and efficiency compared to other methods.

\begin{table}[tb]
  \centering
  \begin{tabular}{@{}lccccc@{}}
    \hline
    Method & Model &PT Ep. & PT Hrs.  & AP$^{b}$ & AP$^{m}$\\
    \hline
     SimMIM$_{192}$& Swin-B & 800 &656 & 50.4& 44.4\\
     GreenMIM& Swin-B & 800 & 296 & 50.0& 44.1\\
     LocalMIM& Swin-B & 400 & 200& 50.7& 44.9\\
     WaMIM& Swin-B & 400 & 160& \textbf{50.9} & \textbf{45.1} \\
  \hline
  \end{tabular}
  \caption{The COCO results for object detection and instance segmentation with metrics including AP$^{b}$ ($\%$) and AP$^{m}$ ($\%$). All models are fine-tuned end-to-end using the Mask R-CNN framework with the pre-trained Swin-B backbone.
  }
  \label{tab:coco}
  \vspace{-0.5em}
\end{table}

\noindent\textbf{Object Detection and Instance Segmentation.} 
We evaluate our method’s transferability to downstream dense tasks, testing its performance on object detection and instance segmentation with the COCO dataset. Table~\ref{tab:coco} shows AP$^b$ (object detection), AP$^m$ (instance segmentation), and total pre-training hours across methods. Our method achieves gains of $0.5$, $0.9$, and $0.2$ in object detection, and $0.7$, $1.0$, and $0.2$ in instance segmentation, compared to SimMIM, GreenMIM, and LocalMIM, respectively, while incurring only $24\%$, $54\%$, and $80\%$ of their computational costs.

\noindent\textbf{Semantic Segmentation.} 
We further evaluate our method on semantic segmentation tasks using the ADE20K dataset, reporting mIoU scores and total pre-training hours for different methods in Table~\ref{tab:ade20k}. The results demonstrate that our method achieves comparable or superior performance while being significantly more efficient.

\begin{table}[tb]
  \centering
  \begin{tabular}{@{}lcccc@{}}
    \hline
    Method& Model & PT Ep. & PT Hrs.  & mIoU\\
    \hline
     MoCo v3& ViT-B & - &- & 47.3\\
     BEiT& ViT-B & 800 &824 & 47.1\\
     MAE& ViT-B &1600  & 752 & 48.1\\
     MaskFeat& ViT-B &1600  &2672 &\textbf{48.8} \\
     CAE& ViT-B & 800 &960 & \textbf{48.8}\\
     LocalMIM& ViT-B & 400 & 120 & 47.2 \\
     WaMIM& ViT-B & 400 & 96 & {48.7} \\
  \hline
  \end{tabular}
  \caption{The ADE20K results for semantic segmentation in terms of mIoU. All models are fine-tuned end-to-end using UperNet with the pre-trained ViT-B backbone.
  }
  \vspace{-0.5em}
  \label{tab:ade20k}
\end{table}

\begin{table*}
  \centering
  \begin{subtable}{0.32\linewidth}
    \centering
    \setlength{\tabcolsep}{6pt}{
    {
    \begin{tabular}{x{28}x{28}x{32}}
    \hline
    ratio & ViT-B & Swin-B  \\
    \hline
    0.4 & 83.1 & 83.6\\
    0.6 & 83.3 & 83.8\\
    0.75 & \underline{\textbf{83.4}} & \underline{\textbf{83.9}} \\
    0.9 & 82.9 & 83.4\\
    \hline
    \end{tabular}
    }
    \caption{}
    \label{tab:mask_ratio}
    }
  \end{subtable}
  \hfill
  \begin{subtable}{0.32\linewidth}
    \centering
    \setlength{\tabcolsep}{6pt}{
    {
    \begin{tabular}{x{32}x{28}x{32}}
        \hline
        levels & ViT-B& Swin-B \\
        \hline
        5 & \underline{\textbf{83.4}} & \underline{\textbf{83.9}} \\
        4 & 83.4 & 83.9 \\
        3 & 83.2 & 83.7 \\
        \hline
    \end{tabular}
    }
    \caption{}
    \label{tab:wavelet_levels}
    }
    \end{subtable}
    \hfill
  \begin{subtable}{0.32\linewidth}
    \centering
    \setlength{\tabcolsep}{6pt}{
    {
    \begin{tabular}{x{50}x{28}x{32}}
        \hline
        decoder & ViT-B& Swin-B \\
        \hline
        512D-16H & 83.4 & 83.8 \\
        256D-8H & \underline{\textbf{83.4}} & 83.9 \\
        128D-4H & {83.2}& \underline{\textbf{83.9}} \\
        \hline
    \end{tabular}
    }
    \caption{}
    \label{tab:decoder_design}
    }
  \end{subtable}
    \hfill
  \begin{subtable}{0.32\linewidth}
    \centering
    \setlength{\tabcolsep}{2pt}{
    {
    \begin{tabular}{y{48}x{48}x{24}}
        \hline
        locations & model & Top-1  \\
        \hline
        $\{12\}$ & \multirow{4}{*}{ViT-B} &82.9 \\
        $\{1, 2, 11, 12\}$ & & {83.2} \\
        $\{2, 4, 10, 12\}$ & &{{83.3}}  \\
        $\{3, 6, 9, 12\}$ & &\underline{\textbf{83.4}}  \\
        \hline
    \end{tabular}
    }
    \caption{}
    \label{tab:locations_vit}
    \vspace{-0.5em}
    }
  \end{subtable}
    \hfill
    \begin{subtable}{0.32\linewidth}
    \centering
    \setlength{\tabcolsep}{2pt}{
    {
    \begin{tabular}{y{48}x{48}x{28}}
    \hline
        locations & model & Top-1  \\
        \hline
        $\{24\}$ & \multirow{4}{*}{Swin-B} &83.2 \\
        $\{22,24\}$ & &{83.5} \\
        $\{4,22,24\}$ & &{{83.8}}  \\
        $\{2,4,22,24\}$ & &\underline{\textbf{83.9}}  \\
        \hline
    \end{tabular}
    }
    \caption{}
    \label{tab:locations_swin}
    \vspace{-0.5em}
    }
  \end{subtable}
    \hfill
  \begin{subtable}{0.32\linewidth}
    \centering
    \setlength{\tabcolsep}{2pt}{
    {
    \begin{tabular}{y{70}x{28}x{32}}
    \hline
    loss weight & ViT-B & Swin-B \\
    \hline
    $\{1,1,1,1\}$ & 83.3 & 83.8 \\
    $\{0.8,0.9,1.1,1.2\}$ & \underline{\textbf{83.4}} & \underline{\textbf{83.9}} \\
    $\{0.7,0.8,1.2,1.3\}$ & {83.2} & 83.7 \\
    $\{1.2,1.1,0.9,0.8\}$ & {83.1} & 83.5 \\
    \hline
    \end{tabular}
    }
    \caption{}
    \label{tab:loss_weight}
    \vspace{-0.5em}
    }
  \end{subtable}
  \vspace{-0.5em}
  \caption{{WaMIM ablations}. (a) \textbf{Mask ratio}. A mask ratio of 0.75 performs best for both ViT-B and Swin-B. (b) \textbf{Wavelet decomposition levels}. Five-level wavelet decomposition demonstrates the best performance and efficiency. (c) \textbf{Decoder design}. The light-weight decoder matches the heavier one in performance with greater efficiency. (d) \textbf{Locations for ViT}. The uniform division for chosen layers achieves the best performance for ViT. (e) Predicting multi-frequency, multi-scale targets boosts performance and efficiency. (f) \textbf{Loss weight}. Emphasizing low-frequency, high-level targets improves representation learning.}
  \label{tab:ablations}
  \vspace{-0.5em}
\end{table*}


\subsection{Ablation studies} 
In this part, we conduct ablation experiments to assess the impact of key components in our approach and validate the design choices. 
Table~\ref{tab:mask_ratio}$\sim$\ref{tab:loss_weight} show the WaMIM ablation experimental results with ViT-B and Swin-B on ImageNet-1K.
All models are pre-trained for 100 epochs. We report fine-tuning Top-1 accuracy (\%). If not specified, the default setting is: the mask ratio is $75\%$, the decoder consists of a single Transformer block with a dimension of 256 and 8 heads for ViT and a dimension of 128 and 4 heads for Swin, the locations for output features are the output of each stage for Swin and the output of $\{3, 6, 9, 12\}$-th layer for ViT, the wavelet decomposition level is 5, the loss weight is $\{0.8,0.9,1.1,1.2\}$. The selected settings are \underline{underlined}.

\noindent \textbf{Mask ratio.} 
We experiment with mask ratios ranging from 0.4 to 0.9 in our method, as shown in Table~\ref{tab:mask_ratio}. The results indicate that increasing the mask ratio from 0.4 to 0.75 improves performance on ViT-B and Swin-B. However, at a high mask ratio of 0.9, performance declines, suggesting that the reconstruction task becomes too difficult for the model to learn. Thus, we selected 0.75 as the default mask ratio.

\noindent \textbf{Decoder design.} 
Numerous studies~\cite{huang2022green,li2022uniform} show that light-weight decoders effectively enable MIM to acquire generalizable representations. In our framework, we create multiple reconstruction targets at different frequencies and scales, requiring several independent decoders. To reduce computational costs, we explore the use of lighter decoders. This section examines how the embedding dimension and the number of self-attention heads in a single-transformer-block decoder affect performance. As shown in Table~\ref{tab:decoder_design}, our approach achieves comparable or better results, even with significantly lighter decoders.

\noindent \textbf{Wavelet decomposition levels.} 
Wavelet decomposition levels determine the frequency and scale granularity represented by each level of wavelet coefficients. For example, a $5$-level decomposition on an image yields coefficients across $6$ frequency bands with corresponding scales of $\{112^2, 56^2, 28^2, 14^2, 7^2\}$. We experiment with $3$, $4$, and $5$ decomposition levels, setting the number of reconstruction targets to $3$, $4$, and $4$, respectively. We then select coefficients from the $\{1, 2, 3\}$-th, $\{1, 2, 3, 4\}$-th, and $\{2, 3, 4, 5\}$-th levels, corresponding to scales of $\{112^2, 56^2, 28^2\}$, $\{112^2, 56^2, 28^2, 14^2\}$, and $\{56^2, 28^2, 14^2, 7^2\}$, ranging from high to low frequency. As shown in Table~\ref{tab:wavelet_levels}, the $5$-level wavelet decomposition offers the best balance of efficiency and performance.

\noindent \textbf{Loss weight.} 
Loss weights determine whether we prioritize reconstructing high-frequency, low-level targets or low-frequency, high-level targets during representation learning. Table~\ref{tab:loss_weight} shows our method's performance under different loss weight configurations. Lower indices in the array correspond to weights for high-frequency, low-level targets, while higher indices represent weights for low-frequency, high-level targets. The results suggest that moderately prioritizing low-frequency, high-level features enhances representation learning. However, overemphasizing them can hinder performance, underscoring the importance of incorporating high-frequency, low-level information in MIM.

\noindent \textbf{Locations for multi-level features.} 
In our framework, the locations of multi-level features used to predict reconstruction targets is crucial because each transformer layer focuses on different image frequencies and scales. Shallow layers primarily capture high-frequency, low-level details, while deep layers capture low-frequency, high-level semantic information. Thus, selecting the appropriate features to match the frequency and scale of reconstruction targets can significantly improve representation learning efficiency and performance.
For the hierarchical Swin architecture, we select output features from its four stages, corresponding to the $\{2, 4, 22, 24\}$-th layers from shallow to deep. In our experiments, we progressively add lower-stage features, starting from the last stage, to demonstrate the benefit of predicting multiple reconstruction targets across different frequencies and scales. As shown in Table~\ref{tab:locations_swin}, predicting multiple carefully designed targets enhances both efficiency and performance.
For the columnar ViT, we test various layer combinations, focusing on $\{1, 2, 11, 12\}$-th (head and tail division), $\{2, 4, 10, 12\}$-th (Swin-style division), and $\{3, 6, 9, 12\}$-th (uniform division). The results in Table~\ref{tab:locations_vit} indicate that uniform division yields the best performance for ViT.

\section{Conclusion}
In this paper, we introduce an efficient approach to enhance masked image modeling (MIM) by leveraging wavelet transform to guide the construction of multi-level reconstruction targets representing different frequencies and scales. We employ multi-level features to predict these reconstruction targets separately. In practice, we directly utilize multi-level wavelet coefficients as reconstruction targets, ensuring that our method is both straightforward and efficient. Moreover, our approach can seamlessly integrate into most existing MIM methods. Extensive experimental results indicate that, compared to other methods, our approach achieves comparable or superior results on numerous downstream tasks, while also demonstrating significantly higher efficiency.

\section{Acknowledgments}
This work was supported in part by the National Natural Science Foundation under Grant 62402251, 62472238.

\bigskip

\bibliography{aaai25}

\clearpage

\begin{appendix}
\noindent \begin{center} {\large  \textbf{Appendix}} \end{center}
\setcounter{table}{4}
\setcounter{figure}{2}

In this appendix, we first demonstrate the experimental results for robustness on several ImageNet OOD datasets in Appendix~\ref{sec:b}. Then we provide the detailed experimental settings for pre-training, image classification on ImageNet-1K~\cite{russakovsky2015imagenet} dataset, object detection and instance segmentation on COCO~\cite{lin2014microsoft} dataset, and semantic segmentation on ADE20k~\cite{zhou2017scene} dataset in Appendix~\ref{sec:a}.

\section{Experimental Results for Robustness}
\label{sec:b}
Performance to out-of-distribution (OOD) dataset is a common evaluation for the robustness of models. We compare the performance of different methods on four ImageNet OOD variants: ImageNet-Corruption~\cite{hendrycks2018benchmarking}, ImageNet-Adversarial~\cite{hendrycks2021natural}, ImageNet-Rendition~\cite{hendrycks2021many}, and ImageNet-Sketch~\cite{wang2019learning}. We report top-1 accuracy on ImageNet-A/R/S and mCE on ImageNet-C. The Average Score is calculated as the average of all results, and (1-mCE) is used for the calculation. 
As shown in Table~\ref{tab:robustness}, our method achieves an Average Score of 43.9, which is 2.1, 1.7, and 0.4 higher than MAE, MFM, and LocalMIM, respectively, while only using 13\%, 29\%, and 80\% of their computational resources. This also indicates that our method demonstrates better robustness.

\begin{table*}[htb]
  \centering
  \begin{tabular}{@{}lcccccccc@{}}
    \hline
    Method & Backbone &PT Epoch & Total PT Hours  & IN-A & IN-R &IN-S&IN-C $\downarrow$& Avg. Score\\
    \hline
     MAE~\cite{he2022masked} & ViT-B & 1600 & 752 & \textbf{35.9} & 48.3 & 34.5 & 51.7 &41.8\\
     MFM~\cite{xie2022masked} & ViT-B & 300 &  330 & 32.7 & 48.6 & 34.8 & \textbf{47.5}& 42.2 \\
     LocalMIM*~\cite{wang2023masked}& ViT-B & 400 & 120 & 34.6 & 50.9 & 37.0 & 48.5 & 43.5\\
     WaMIM& ViT-B & 400 & 96 & 35.4 & \textbf{51.5} & \textbf{37.5} & 48.1 & \textbf{43.9}\\
  \hline
  \end{tabular}
  \caption{Robustness evaluation on ImageNet OOD variants. We report Top-1 accuracy for ViT-B, except for ImageNet-C, where we use the mean corruption error (mCE). And (1-mCE) is used for the calculation of the Average Score. * indicate that the results are reproduced with the official code.
  }
  \label{tab:robustness}
\end{table*}

\section{Detailed Experiment Settings}

\label{sec:a}

\subsection{Pretraining}

The default configurations are displayed in Table~\ref{tab:impl_pretraining}. Specifically, we utilize the ViT-S~\cite{dosovitskiy2021an}, ViT-B~\cite{dosovitskiy2021an} and Swin-B~\cite{liu2021swin} models in our experiments. These models undergo pre-training for 100/300/400 epochs with warmup epochs set at 10/30/40, respectively, while maintaining a total batch size of 2048. We employ the AdamW optimizer~\cite{loshchilov2018decoupled} with a weight decay of 0.05 and $\beta_1 = 0.9$, $\beta_2 = 0.95$. The base learning rate is initialized as 2$e-$4 for ViT and 1$e-$4 for Swin, with a cosine learning rate schedule~\cite{loshchilov2017sgdr}. Additionally, the effective learning rate is linearly scaled: $lr = base\_lr \times batch\_size / 256$. For the wavelet settings, we set the wavelet decomposition levels to $5$, and the selected coefficients are $\{1, 2, 3, 4\}$-th levels. The locations for multi-level features are $\{3,6,9,12\}$ for ViT and $\{2,4,22,24\}$ for Swin, respectively, with a loss weight set to $\{0.8,0.9,1.1,1.2\}$. Simple data augmentations, such as random cropping and horizontal flip, are applied, and the mask ratio is set to 75\%. Finally, we initialize all Transformer blocks using Xavier uniform initialization~\cite{glorot2010understanding}, following the  MAE~\cite{he2022masked}.

\begin{table}[htp]
\centering
\begin{tabular}{y{95}|x{110}}
\shline
config & ViT-S, ViT-B; Swin-B \\
\shline
optimizer & AdamW \\
base learning rate & 2e-4; 1e-4 \\
weight decay & 0.05 \\
optimizer momentum & $\beta_1, \beta_2{=}$ 0.9, 0.95 \\
batch size & 2048 \\
learning rate schedule & cosine decay \\
pre-training epochs & 100/300, 100/400; 100/400 \\
warmup epochs & 10/30, 10/40; 10/40 \\
augmentation & random cropping\& \ \ \ \  horizontal flip \\
mask ratio &  75\%  \\
wavelet decomp. levels & $5$ \\
selected coefficients & $\{1,2,3,4\}$ \\
locations & $\{2,4,22,24\}; \{3,6,9,12\}$ \\
loss weight & $\{0.8,0.9,1.1,1.2\}$ \\
pre-training resolution & 224 $\times$ 224 \\
\shline
\end{tabular}
\vspace{.5em}
\caption{{Pre-training settings.}}
\label{tab:impl_pretraining} 
\end{table}

\subsection{Image Classification}
\begin{table}[h]
\centering
\begin{tabular}{y{106}|x{98}}
\shline
config & ViT-S, ViT-B; Swin-B \\
\shline
optimizer & AdamW \\
base learning rate & 4e-4 \\
weight decay & 0.05 \\
optimizer momentum & $\beta_1, \beta_2{=}$ 0.9, 0.999 \\
layer-wise lr decay & 0.8, 0.75; 0.8 \\
batch size & 2048 \\
learning rate schedule & cosine decay \\
training epochs & 200, 100; 100 \\
warmup epochs & 20 \\
augmentation & RandAug (9, 0.5) \\
label smoothing & 0.1 \\
mixup & 0.8 \\
cutmix  & 1.0 \\
drop path rate & 0.1 \\
fine-tuning resolution & 224 $\times$ 224 \\
\shline
\end{tabular}
\vspace{.5em}
\caption{{Fine-tuning settings for image classification.}}
\label{tab:impl_finetuning} 
\end{table}

Table~\ref{tab:impl_finetuning} outlines the default configuration. Training for ViT-S ViT-B, and Swin-B spans 200, 100 and 100 epochs, with 20 warmup epochs. The base learning rate is set to 4$e-$3 for all models, and the layer-wise learning rate decay is adjusted to 0.8, 0.75, 0.8 for ViT-S, ViT-B and Swin-B, respectively. The drop path rate~\cite{huang2016deep} is fixed at 0.1, while the batch size remains constant at 2048. During training, robust data augmentation techniques are employed, including label smoothing~\cite{szegedy2016rethinking}, mixup~\cite{zhang2017mixup}, cutmix~\cite{yun2019cutmix}, and randAugment~\cite{cubuk2020randaugment}. Notably, global pooling features replace class tokens during fine-tuning for ViT. Fine-tuning involves all models being trained using an image resolution of 224 $\times$ 224, with adherence to the linear $lr$ scaling rule: $lr = base\_lr \times batch\_size /$256.

\subsection{Object Detection and Instance Segmentation}
The default setup is illustrated in Table~\ref{tab:impl_coco}. To incorporate the pre-trained Swin-B model into Mask R-CNN~\cite{he2017mask}, we tailor it for compatibility with an FPN backbone~\cite{lin2017feature}. We deploy multi-scale training, where images are resized to have a short side ranging between 480 and 800 pixels and a long side no greater than 1333 pixels. For optimization, we employ the AdamW optimizer with a learning rate set to 1$e-$4, a weight decay of 0.05. And the total batch size is 16. The drop path rate is fixed at 0.3. We adhere to a 3$\times$ training schedule, encompassing 36 epochs with the learning rate decayed by 10$\times$ at epochs 27 and 33.

\begin{table}[h]
\centering
\begin{tabular}{y{106}|x{98}}
\shline
config & Swin-B \\
\shline
optimizer & AdamW \\
base learning rate & 1e-4 \\
weight decay & 0.05 \\
optimizer momentum & $\beta_1, \beta_2{=}$ 0.9, 0.999 \\
batch size & 16 \\
learning rate schedule & step decay \\
training epochs & 36 \\
drop path & 0.3 \\
\shline
\end{tabular}
\vspace{.5em}
\caption{{Fine-tuning settings for object detection and instance segmentation.}}
\label{tab:impl_coco} 
\end{table}

\subsection{Semantic Segmentation}
The default configuration is depicted in Table~\ref{tab:impl_ade20k}. We utilize the pre-trained ViT-B as the backbone and integrating UperNet~\cite{xiao2018unified} as our segmentation model. Our fine-tuning process involves $160k$ iterations using the AdamW optimizer, with a base learning rate set to 4e-4, weight decay at 0.05, and a batch size of 16. The learning rate warms up for 1500 iterations before undergoing linear decay strategy. And the input resolution remains fixed at $512\times512$.

\begin{table}[h]
\centering
\begin{tabular}{y{106}|x{98}}
\shline
config & ViT-B \\
\shline
optimizer & AdamW \\
base learning rate & 4e-4 \\
weight decay & 0.05 \\
optimizer momentum & $\beta_1, \beta_2{=}$ 0.9, 0.999 \\
batch size & 16 \\
learning rate schedule & linear decay \\
training iterations & $160k$ \\
drop path & 0.1 \\
\shline
\end{tabular}
\vspace{.5em}
\caption{{Fine-tuning settings for semantic segmentation.}}
\label{tab:impl_ade20k} 
\end{table}

\end{appendix}

\end{document}